\documentclass[10pt,twocolumn,letterpaper]{article}

\usepackage{iccv}      

\definecolor{iccvblue}{rgb}{0.21,0.49,0.74}
\usepackage[pagebackref,breaklinks,colorlinks,allcolors=iccvblue]{hyperref}

\usepackage{pifont}
\usepackage{makecell}
\usepackage{multirow}
\usepackage{arydshln} 
\usepackage{graphicx}
\usepackage{booktabs}
\usepackage{array}
\usepackage{caption}
\usepackage{colortbl}
\usepackage{xparse}
\usepackage{pifont} 

\ExplSyntaxOn
\NewExpandableDocumentCommand { \cdashedline } { }
  {
    \noalign { \skip_vertical:n { 0.3 mm } }
    \__johan_cline_i:nn 1
  }
\cs_set:Npn \__johan_cline_i:nn #1 #2 { \__johan_cline_i:w #1-#2 \q_stop }
\cs_set:Npn \__johan_cline_i:w #1-#2-#3 \q_stop
  {
    \int_compare:nNnT { #1 } < { #2 }
      { \multispan { \int_eval:n { #2 - #1 } } & }
    \multispan { \int_eval:n { #3 - #2 + 1 } }
      {
        \skip_horizontal:N \tabcolsep
        \xleaders
        \hbox
          {
            \vrule height 0.4pt depth 0pt width 2pt
            \kern 1pt
          }
        \hfill
        \vrule height 0.4pt depth 0pt width 2pt
        \skip_horizontal:N \tabcolsep
      }
    \peek_meaning_remove_ignore_spaces:NTF \cdashedline
      { & \exp_args:Ne \__johan_cline_i:nn { \int_eval:n { #3 + 1 } } }
      {
        \everycr { } \cr
        \noalign { \skip_vertical:n { 0.3 mm } }
      }
  }
\ExplSyntaxOff
\title{ E-SAM: Training-Free Segment Every Entity Model }

\author{Weiming Zhang$^{1}$ \quad Dingwen Xiao$^{1}$ \quad Lei Chen$^{1,2}$ \quad Lin Wang$^{3\dag}$
\vspace{1mm}\\
$^{1}$ HKUST (GZ) \quad $^{2}$ HKUST \quad $^{3}$ Nanyang Technological University
\vspace{1mm}\\}

\begin{document}

\twocolumn[{
\renewcommand\twocolumn[1][t!]{#1}%
\maketitle

\begin{center}
    \centering
    \vspace{-10pt}
    \includegraphics[width=0.95\textwidth]{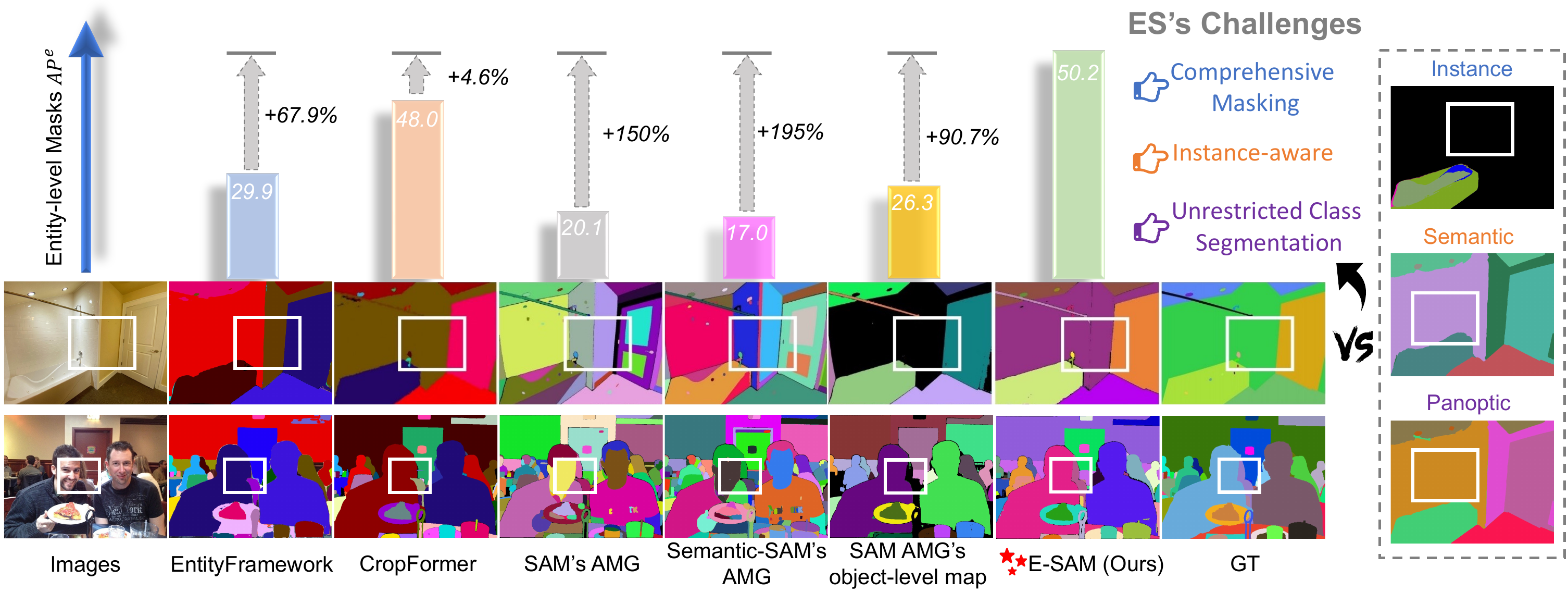}
    \vspace{-8pt}
    {\captionof{figure}{
    Compared with prior ES methods~\cite{qi2022open,qi2022high}, SAM and Semantic-SAM~\cite{li2023semantic}, our E-SAM demonstrates superior performance. Notably, our E-SAM, without any additional training, significantly outperforms both SAM's AMG and its object-level map.}}
    \label{coverfigure}
\end{center}}]

\begin{abstract}

Entity Segmentation (ES) aims at identifying and segmenting distinct entities within an image without the need for predefined class labels. This characteristic makes ES well-suited to open-world applications with adaptation to diverse and dynamically changing environments, where new and previously unseen entities may appear frequently.
Existing ES methods either require large annotated datasets or high training costs, limiting their scalability and adaptability.
Recently, the Segment Anything Model (SAM), especially in its Automatic Mask Generation (AMG) mode, has shown potential for holistic image segmentation. However, it struggles with over-segmentation and under-segmentation, making it less effective for ES.
In this paper, we introduce \textbf{E-SAM,} a novel training-free framework that exhibits exceptional ES capability.
Specifically, we first propose \textbf{Multi-level Mask Generation (MMG)} that
hierarchically processes SAM's AMG outputs to generate reliable object-level masks while preserving fine details at other levels. \textbf{Entity-level Mask Refinement (EMR)} then
refines these object-level masks into accurate entity-level masks. That is, it separates overlapping masks to address the redundancy issues inherent in SAM's outputs and merges similar masks by evaluating entity-level consistency. Lastly, \textbf{Under-Segmentation Refinement (USR)} addresses under-segmentation by generating additional high-confidence masks fused with EMR outputs to produce the final ES map. These three modules are seamlessly optimized to achieve the best ES without additional training overhead.
Extensive experiments demonstrate that E-SAM achieves state-of-the-art performance compared to prior ES methods, demonstrating a significant improvement by \textbf{+30.1} on benchmark metrics.
Code is available at \url{https://}.

\end{abstract}

\vspace{-10pt}
\section{Introduction}
\begin{figure}[t!]
    \centering
    \includegraphics[width=0.49\textwidth]{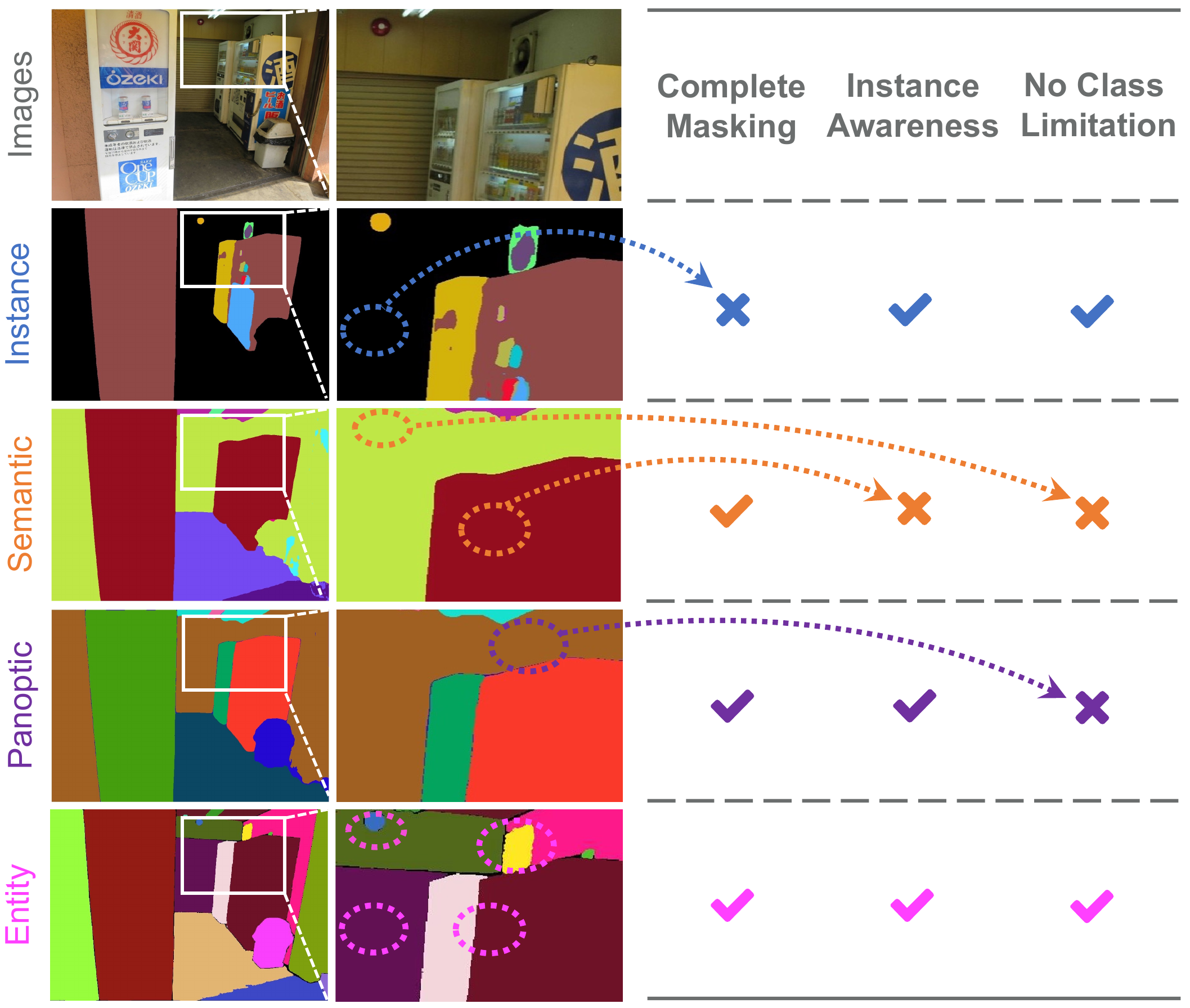}
    \vspace{-10pt}
    \caption{Comparison of ES and with three segmentation tasks.
    }
    \vspace{-15pt}
    \label{Fig: difftask}
\end{figure}

  \vspace{-5pt}
Entity Segmentation (ES) \cite{qi2022open} is an emerging task in computer vision that focuses on segmenting visual entities in an image without relying on predefined class labels. Unlike traditional segmentation tasks, which are limited by fixed categories, ES aligns more closely with human perception \cite{man1982computational}, where entities are identified based on visual coherence.
As shown in Figs. \textcolor{iccvblue}{1} \& \ref{Fig: difftask}, ES
offers \textit{comprehensive masking} in instance segmentation, enhances the \textit{instance awareness} of semantic segmentation, and overcomes the \textit{label class limitations} in panoptic segmentation, where certain undefined entities (e.g., the metal rod in Fig. \textcolor{iccvblue}{1} and exhaust fan in Fig.~\ref{Fig: difftask}) are failed to be segmented.
This class-agnostic nature makes ES well-suited to various open-world applications, from image editing \cite{li2024mike} and manipulation \cite{wang2022manitrans, wang2023entity} to real-time environments such as autonomous driving \cite{feng2020deep, yuan2024graph}, robotics \cite{cao2024adapting, xie2021unseen, durner2021unknown}, and surveillance \cite{hampapur2003smart, gruosso2021human}.
However, existing ES methods \cite{qi2022open,qi2022high,cao2024sohes,wang2024segment,qi2024unigs} often face high training costs and require extensive annotated datasets, limiting their scalability and practicality. In addition, these models struggle with generalization in diverse scenarios.

The Segment Anything Model (SAM) ~\cite{kirillov2023segment}, as a foundational model for segmentation, demonstrates robust zero-shot performance due to its training on a vast dataset containing over one billion masks.
SAM can accept various prompts, such as points, boxes, and masks, to segment target entities and allows for the continuous refinement of segmentation results through additional prompts.
Furthermore, SAM's Automatic Mask Generation (AMG) mode was specifically designed to perform full-image instance segmentation without the need for explicit prompts~\cite{xu2023eviprompt, shu2023tinysam, li2023semantic}.
However, due to AMG's inherent limitations, its performance for ES is suboptimal.
Specifically, the uniform point prompt sampling strategy in AMG enables SAM to generate three mask levels (object, part, sub-part) for each point. However, applying a naive Non-Maximum Suppression (NMS) \cite{neubeck2006efficient} to remove redundant masks often results in over-segmentation and under-segmentation, either discarding fine details or failing to remove unnecessary overlaps,
see Fig. \textcolor{iccvblue}{1}.
To overcome these limitations, we explore a novel question: \textit{How can we efficiently and effectively achieve ES of all entities in an image?}

To this end, we propose \textbf{E-SAM}, a novel \textit{training-free} framework specifically designed the achieve state-of-the-art ES performance without incurring additional training costs. Our E-SAM effectively mitigates the over-segmentation and under-segmentation challenges inherent in AMG by integrating three key modules: Multi-level Mask Generation (MMG) (Sec.\ref{Sec: MMG}), Entity-level Mask Refinement (EMR) (Sec.\ref{Sec: EMR}), and Under-Segmentation Refinement (USR) (Sec.\ref{Sec: USR}).
Specifically, the MMG module first categorizes the AMG's outputs according to their area levels and confidence scores, subsequently applying diverse NMS strategies with varying thresholds to retain high-confidence object-level masks while preserving additional masks at the part and subpart levels in densely populated regions.
Then, the EMR module begins by increasing the number of uniformly sampled points to construct a mask gallery, which includes a diverse set of high-confidence masks. Next, this gallery is utilized to identify and separate overlapping object-level masks, refining them into distinct adjacent masks to eliminate redundancy. Subsequently, EMR constructs a similarity matrix among these refined masks and leverages the mask gallery to merge highly similar masks, ultimately producing an accurate entity-level segmentation map.
Lastly, the USR module refines under-segmented regions in EMR's outputs by incorporating superpixel centroids, as well as part and subpart mask centroids, as prompts to guide additional segmentation. By seamlessly integrating the three modules, our E-SAM efficiently generates high-quality ES masks for the entire image without incurring additional training overhead.

We conducted extensive experiments that demonstrated the effectiveness of our framework across multiple datasets. As shown in Fig. \textcolor{iccvblue}{1}, our E-SAM consistently outperforms the previous state-of-the-art ES methods \cite{qi2022open,qi2022high} and SAM. In particular, under the same backbone size, our E-SAM consistently outperformed SAM's AMG by more than double in performance according to benchmark metrics. As illustrated in Fig. \ref{Fig: Robust_Comparsion}, our E-SAM exhibits strong generalization capability even in unseen datasets (or open-world scenarios), underscoring the novelty and effectiveness of our E-SAM design, particularly in its training-free approach.

\begin{figure*}[t!]
    \centering
    \includegraphics[width=0.90\textwidth]{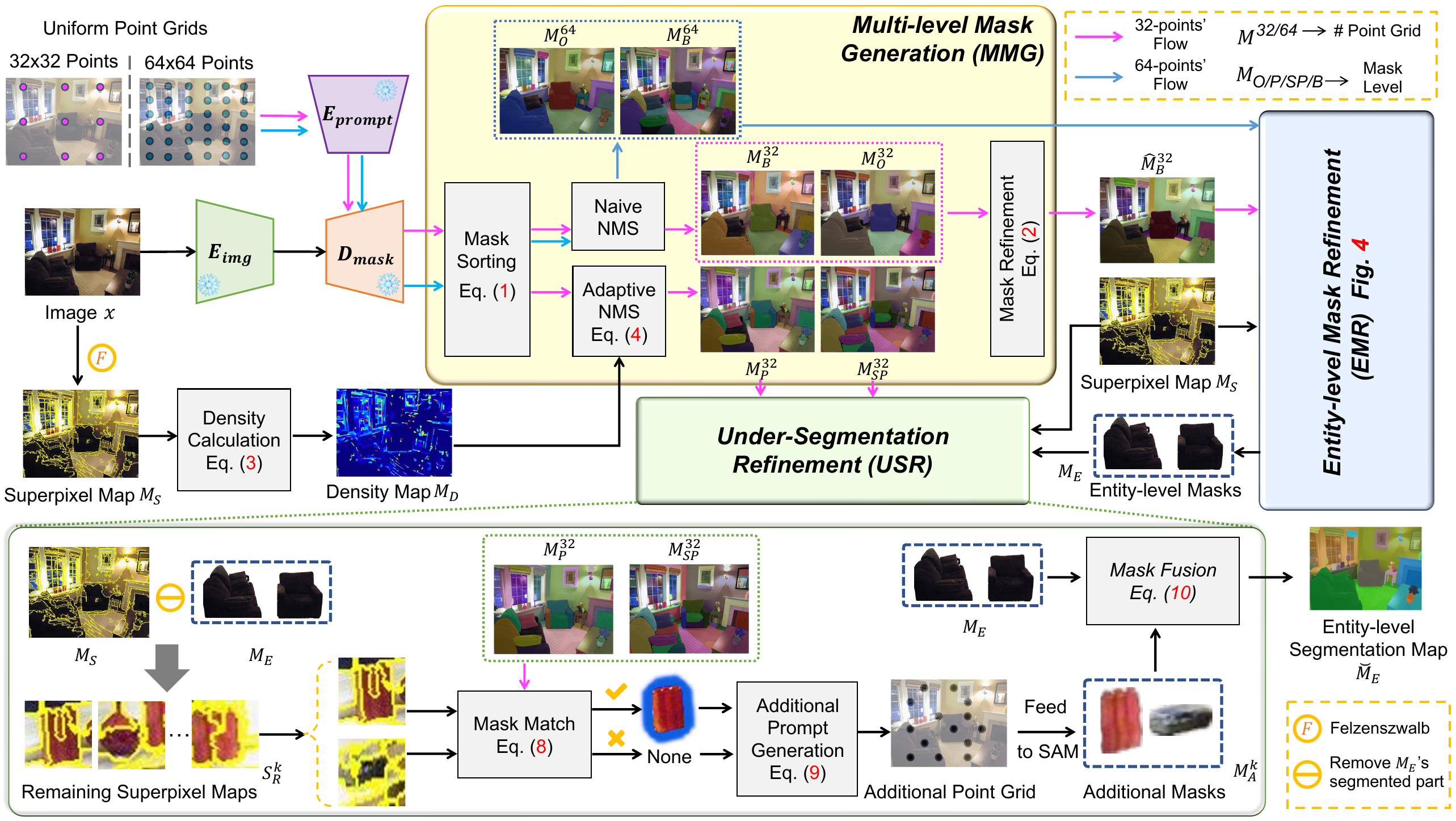}
    \vspace{-5pt}
    \caption{Overview of our E-SAM framework. Our method contains three main technical components:  Multi-level Mask Generation (MMG), Entity-level Mask Refinement (EMR), and Under-Segmentation Refinement (USR).
    }
    \vspace{-15pt}
    \label{Fig: framework}
\end{figure*}

In summary, our contributions are as follows: \textbf{(I)} We introduce E-SAM, a novel \textbf{training-free} framework aimed at enhancing SAM's performance in entity segmentation. \textbf{(II)} We design three modules, MMG, EMR, and USR, that work in sequence to generate reliable masks at multiple levels, eliminate overlapping regions, refine masks at the entity level, and optimize undersegmented areas, ultimately producing a high-quality entity-level segmentation map. \textbf{(III)} Extensive experiments demonstrate that E-SAM significantly outperforms both SAM and existing ES methods, achieving superior performance across diverse datasets with strong quantitative and qualitative results.

\section{Related Works}
\noindent \textbf{Segment Anything Model (SAM) and AMG.}
SAM \cite{kirillov2023segment} is a foundation model trained on a large-scale dataset containing approximately 1 billion masks, providing robust zero-shot capabilities in segmentation tasks. SAM accepts visual prompts such as points, bounding boxes, or masks, enabling interactive segmentation. This versatility has allowed SAM to find applications in a wide range of domains, including object tracking \cite{cheng2023segment, rajivc2023segment, yang2023track,zhu2023tracking}, 3D instance segmentation \cite{yin2024sai3d,cen2023segment,yang2023sam3d}, and medical imaging \cite{huang2024segment,mazurowski2023segment,zhang2023customized,ma2024segment}.
In addition to its interactive capabilities, SAM includes an Automatic Mask Generation (AMG) mode, which is designed to segment everything within a given image without requiring explicit prompts. SAM's AMG has been explored extensively in recent studies \cite{xu2023eviprompt,kweon2024sam,shu2023tinysam,khan2024segment,guo2023sam} to further extend the capabilities of SAM in various domains.
Despite the successes of SAM's AMG, it is not without challenges. The uniform point sampling strategy in AMG returns three masks per point, often leading to redundancy and overlapping regions. Using a basic Non-Maximum Suppression (NMS)~\cite{neubeck2006efficient} to filter these masks may inadvertently remove crucial object parts(leading to under-segmentation) or retain too many similar masks (causing over-segmentation).
\textit{To address these shortcomings, we propose three modules—MMG, EMR, and USR—that mitigate the issues of over-segmentation and under-segmentation, enabling efficient and accurate segmentation of each entity.}

\noindent \textbf{Entity Segmentation.}
Entity Segmentation (ES) is a novel task that was first introduced in \cite{qi2022open}. The core objective of ES is to segment all perceptually distinct entities within an image, irrespective of any predefined class labels.
Several approaches~\cite{qi2022open,qi2022high,cao2024sohes,wang2024segment,qi2024unigs} have been proposed to solve ES, including methods like CropFormer \cite{qi2022high}, which introduced the EntitySeg dataset with high-quality, densely annotated masks to enhance generalization for segmentation models. However, these specialized models have certain weaknesses. These methods rely heavily on costly, labor-intensive annotated datasets, limiting their scalability and applicability, and hindering real-world generalization. Additionally, training these models requires significant computational resources, adding complexity and limiting broader adoption.
\textit{To address these challenges, our E-SAM optimizes SAM's AMG through a training-free approach to efficiently enhance its performance and achieve high-quality entity-level segmentation maps.}

\section{Methodology}

\subsection{Overview}
In this section, we first provide an overview of our framework -- \textbf{E-SAM}, shown in Fig.~\ref{Fig: framework}. E-SAM operates training-free, eliminating the need for training data or training costs. Therefore, we freeze SAM’s image encoder $E_{\text{img}}$, prompt encoder $E_{\text{prompt}}$, and mask decoder $D_{\text{mask}}$.
Given a test image \(x \in \mathbb{R}^{H \times W \times 3}\), E-SAM follows the SAM's AMG processes~\cite{kirillov2023segment} by inputting the image \(x\) into the image encoder \(E_{\text{img}}\). E-SAM then uniformly generates point prompts along each side, which are fed into the prompt encoder \(E_{\text{prompt}}\).
In the subsequent mask decoding stage \(D_{\text{mask}}\), E-SAM generates and selects high-confidence segmentation masks across the levels of the object, part, and subpart for each point prompt.
Since E-SAM generates masks of multiple granularities, the main challenges of E-SAM lie in: (1) effectively removing overlapping masks, (2) fusing SAM-generated masks to obtain precise entity-level masks, and (3) refining under-segmented regions to ensure complete segmentation of all entities.
To address these challenges, our E-SAM consists of three key modules: Multi-level Mask Generation (MMG) Module (Sec.~\ref{Sec: MMG}), Entity-level Mask Refinement (EMR) Module (Sec.~\ref{Sec: EMR}), and Under-Segmentation Refinement (USR) Module (Sec.~\ref{Sec: USR}). We now describe these modules in detail.

\subsection{Multi-level Mask Generation (MMG)}
\label{Sec: MMG}

The MMG module aims to stratify the outputs of SAM into different levels by employing a combination of NMS methods and thresholds, ultimately generating distinct mask maps to mitigate the over-segmentation issues caused by multi-granularity masks. First, for a given image $x$, MMG uniformly generates 32-point prompts per side. Then, MMG categorizes the masks returned by SAM for each point prompt based on their area into three levels: object-level mask map $M_O^{32}$, part-level mask map $M_P^{32}$, and subpart-level mask map $M_{SP}^{32}$. Additionally, MMG selects the mask with the highest confidence score for each point to form the best-level mask map $M_B^{32}$, as shown in Eq.~(\ref{eq:1}):
\begin{equation}
\small
\label{eq:1}
    \begin{split}
        M_{i}^{32} &= \{M_{i,O}^{32}, M_{i,P}^{32}, M_{i,SP}^{32}\}, \quad A_{i,O} \geq A_{i,P} \geq A_{i,SP}, \\
        \varepsilon_{i,B} &= \mathop{\arg\max}_{\varepsilon \in \{O, P, SP\}} s_{i,\varepsilon}, \quad M_{i,B}^{32} = M_{i,\varepsilon_{i,B}}^{32},
    \end{split}
\end{equation}
where \(i\) denotes the a point prompt, \(A_{i,\varepsilon}\) represents the area and \(s_{i,\varepsilon}\) indicates the confidence score of the \(\varepsilon\)-th level mask returned for point \(i\).
Furthermore, for each mask level, MMG applies specific strategies:
\textbf{Object-Level Mask Map:}
Given that $M_{i,O}^{32}$ best approximates entity-level masks, MMG retains only high-confidence object-level masks, thereby reducing fragile noise in SAM's object-level predictions.
To achieve this, MMG first applies naive NMS with a high threshold, denoted as \(\theta_O\). The remaining object-level masks $M_{i,O}^{32}$ are then checked for presence within the best-level mask map ${M}_B^{32}$ based on Intersection over Union (IoU) criteria. Masks that do not sufficiently match are discarded, resulting in the refined object-level mask map $\hat{M}_{O}^{32}$.
\begin{equation}
\small
    \max \text{ IoU}(M_{i,O}^{32}, M_{*,B}^{32}) \geq \gamma_O
    \implies M_{i,O}^{32} \in \hat{M}_{O}^{32},
\end{equation}
where $\gamma_O$ denotes the IoU threshold.
\textbf{Part-Level and Subpart-Level Mask Maps:}
MMG aims to generate additional masks in complex regions, capturing finer details of densely populated entities.
To facilitate this, MMG first employs Felzenszwalb's method~\cite{Felzenszwalb2004EfficientGI} for superpixel clustering on \(x\), which generates the superpixel map $M_S$, with superpixels denoted as $S_k$. Then, 
MMG creates a density map $M_D$ for \(x\), indicating entity density across regions.
\begin{equation}
\small
    M_D(x) = \sum_{k=1}^{K} w_k \cdot \mathbf{1}_{S_k}(x),
\end{equation}
where $K$ denotes the number of superpixels and $w_k$ denotes the weight assigned to the superpixel.  $\mathbf{1}_{S_k}(x)$ is an indicator function whether pixel belongs to $S_k$.
MMG then uses \(M_D\) in conjunction with Adaptive NMS~\cite{liu2019adaptive} to dynamically adjust the threshold \(N_{\mathcal{M}}\) in densely populated regions.
\begin{equation}
\small
    N_{\mathcal{M}} := \max(N_t, M_D(\mathcal{M})),
\end{equation}
where \(N_t\) is a base threshold and $\mathcal{M}$ denotes the candidate mask.
Given SAM’s robust zero-shot capabilities, precise point prompts allow SAM to provide satisfactory masks for target entities. Thus, MMG also increases the number of point prompts per side to 64 and re-feeds these to \(E_{\text{prompt}}\) to obtain a richer object-level mask map $M_O^{64}$ and a best-level mask map $M_B^{64}$ for subsequent refinement.
\begin{figure}[t!]
    \centering
    \includegraphics[width=0.45\textwidth]{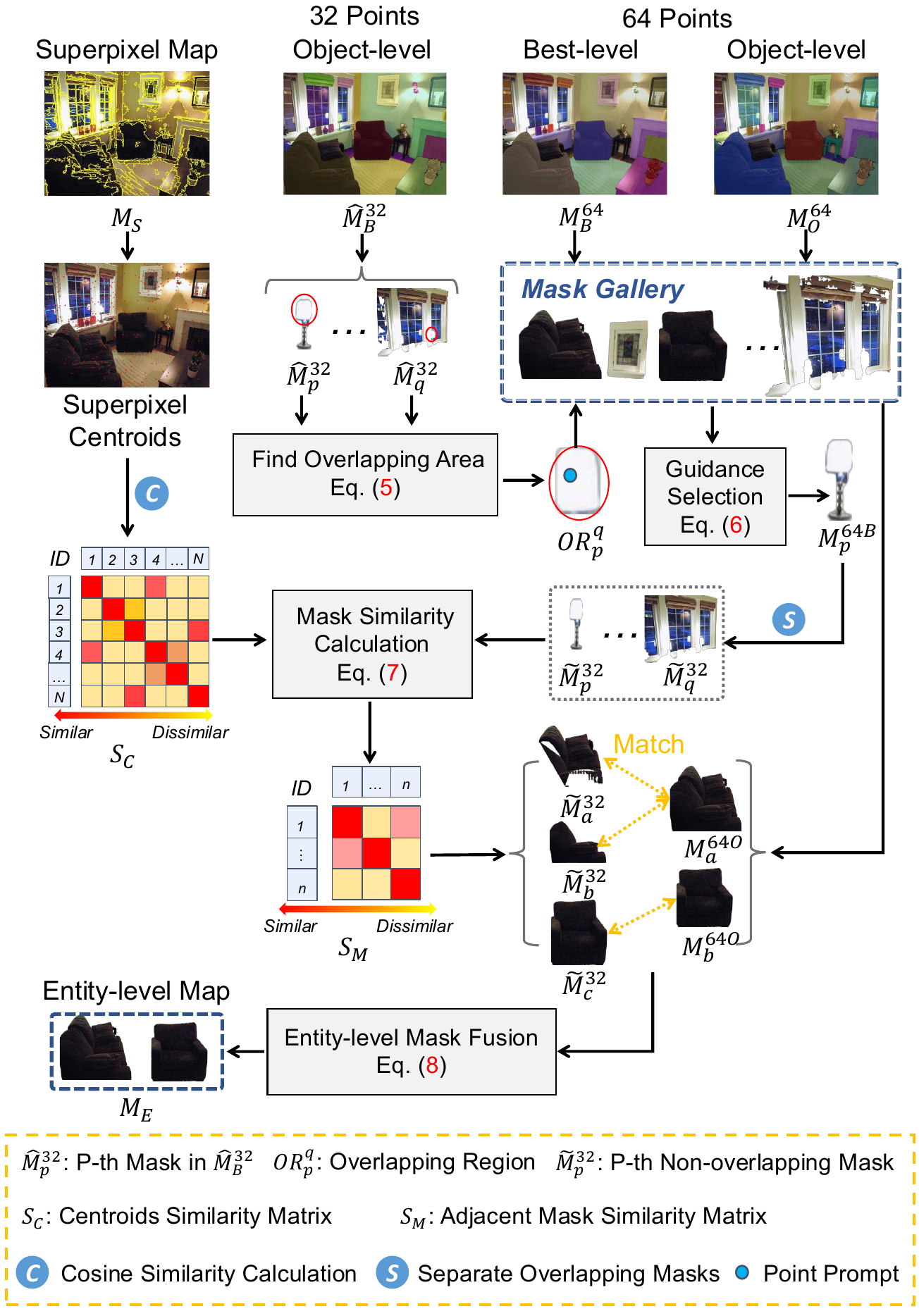}
    \vspace{-8pt}
    \caption{Overview of our EMR framework.
    }
    \vspace{-15pt}
    \label{Fig: EMR}
\end{figure}

\subsection{Entity-level Mask Refinement (EMR)}
\label{Sec: EMR}

The EMR module, illustrated in Fig.~\ref{Fig: EMR}, aims to refine \(\hat{M}_{O}^{32}\) into the entity-level mask map \(M_E\) using the outputs of MMG: \(M_O^{64}\) and \(M_B^{64}\). Specifically, EMR aggregates \(M_O^{64}\) and \(M_B^{64}\) into a mask gallery $G$, which acts as a reference for subsequent refinement.
For the masks in \(\hat{M}_O^{32}\), we employ a ``split-then-merge" strategy to achieve refinement. First, all masks are sorted in descending order based on their predicted scores. For each mask in this sorted list, denoted as \(\hat{M}_p^{32}\), EMR identifies overlapping masks \(\hat{M}_q^{32}\) (where \(p \neq q\)) and calculates the overlap region \(OR_p^q\).
\begin{equation}
\small
    OR_p^q = M_p^{32} \cap M_q^{32}, \quad \forall p \neq q.
\end{equation}
If the area of \(OR_p^q\) relative to the largest mask between \(\hat{M}_p^{32}\) and \(\hat{M}_q^{32}\) is below a threshold \(\delta\), the overlapping region in the larger mask is removed. Otherwise, EMR finds the existing point prompts from the 64-points-per-side sampling strategy within \(OR_p^q\), denoted as \(P^{64}\).
Next, the EMR module considers the masks in the mask gallery corresponding to \(P^{64}\): the object-level mask \(M_p^{64O}\) and the best-level mask \(M_p^{64B}\). We set a tolerance level \(\tau\): if the score difference between \(M_p^{64O}\) and \(M_p^{64B}\) is within \(\tau\), \(M_p^{64O}\) is selected as the guidance for refining the overlap region. Otherwise, \(M_p^{64B}\) is used. When multiple prompts are present, the mask that appears most frequently is chosen for guidance.

\begin{equation}
\small
\label{Eq:6}
    G_p =
    \begin{cases}
        M_p^{64O}, & \text{if } S_p^{64B} - S_p^{64O} < \tau, \\
        M_p^{64B}, & \text{if } S_p^{64B} - S_p^{64O} \geq \tau.
    \end{cases}
\end{equation}
Subsequently, using the selected guidance mask \(G_p\), EMR updates the \(\hat{M}_p^{32}\) and \(\hat{M}_q^{32}\), resulting in non-overlapping masks \(\tilde{M}_p^{32}\) and \(\tilde{M}_q^{32}\). Next, based on the image features extracted by \(E_{\text{img}}\), we construct a centroids cosine similarity matrix \(S_C\) for the centroids in the \(M_S\).
Then, the EMR module identifies the superpixel centroids present within each mask and determines which masks their most similar corresponding centroids belong to. By analyzing the frequency with which each mask appears, we construct the adjacent mask similarity matrix \(S_M\).
\begin{equation}
\small
\vspace{-5pt}
\begin{split}
    S_M(i, j) &= \frac{1}{|C_i|} \sum_{c_i \in C_i} \left | \left \{  c_j \mid c_j \in C_j \wedge c_j \in \text{Top}_k(S_C(c_i, \cdot)) \right \}  \right | ,
\end{split}
\vspace{-8pt}
\end{equation}
where $C_i$ and $C_j$ denotes the set of superpixel centroids in mask \(i\) and \(j\), and $|\cdot|$ marks the cardinality. \(\text{Top}_k(S_C(c_i, \cdot))\) represents the top \(k\) most similar centroids to centroid \(c_i\), based on similarity scores from \(S_C\).
Subsequently, the EMR module evaluates the masks with high similarity scores. It checks the Mask Gallery $G$ to determine if there exists a mask that encompasses both of the highly similar masks. When masks match, they are merged into a unified entity-level mask; otherwise, they remain separate.
\begin{equation}
\small
    {M}_E =
    \begin{cases}
      M_a^{64} \cup M_b^{64}, & \text{if } M_a^{64}, M_b^{64} \in G \text{ and match exists} \\
      \{M_a^{32}, M_b^{32}\}, & \text{otherwise}
    \end{cases}
\end{equation}
After matching and fusing adjacent masks, the EMR produces the entity-level map \(M_E\).

\subsection{Under-Segmentation Refinement (USR)}
\label{Sec: USR}

The USR module, as illustrated in Fig.~\ref{Fig: framework}, aims to address the under-segmentation issues that may arise from the initial outputs of MMG. The module further refines \(M_E\) to ensure comprehensive coverage of all perceptually distinct entities.
The USR first considers the regions in the superpixel map \(M_S\) that are not covered by the entity-level map \(M_E\) and divides them into the remaining superpixel maps \(S_R^k\), where \(k\) represents the number of maps. Then, for each \(S_R^i\), different cases are considered.
When \(S_R^i\) is contained within a mask from \(M_P^{32}\) or \(M_{SP}^{32}\), its centroid is used as an additional point prompt $P_A^i$ for SAM; otherwise, the superpixel's centroid is used. The resulting additional masks generated by USR are denoted as \(M_A^k\).
\begin{equation}
\small
    P_A^i =
    \begin{cases}
        \text{centroid}(M_P^{32} \cup M_{SP}^{32}), & \text{if } S_R^i \subseteq (M_P^{32} \cup M_{SP}^{32},) \\
        \text{centroid}(S_R^i), & \text{otherwise.}
    \end{cases}
\end{equation}
Next, the USR module evaluates each mask in \(M_A^k\) against the current entity-level map \(M_E\). If the IoU between a mask in \(M_A^k\) and an entity-level mask in \(M_E\) exceeds a specified threshold \(\rho\), the mask is retained as part of the entity. Otherwise, it is preserved as an independent entity-level mask. Consequently, \(M_E\) is refined to $\breve{M}_{E}$.
\begin{equation}
\small
    {\breve{M}_{E}}^b =
    \begin{cases}
        M_E^a \cup M_A^a, & \text{if } {iou(M_E^a , M_A^a)} > \rho, \\
        M_A^b, & \text{otherwise.}
    \end{cases}
\end{equation}
To prevent the generation of an excessive number of part-level masks, a naive greedy algorithm is designed within the USR module. This means that USR attempts to use fewer masks from \(M_A^k\) to fill the unsegmented regions in \(M_E\), optimizing the overall segmentation quality.
Through these refinements, USR significantly enhances the robustness and completeness of the segmentation, particularly in regions with complex boundaries or dense entity arrangements.

\begin{figure*}[t!]
    \centering
    \includegraphics[width=0.95\textwidth]{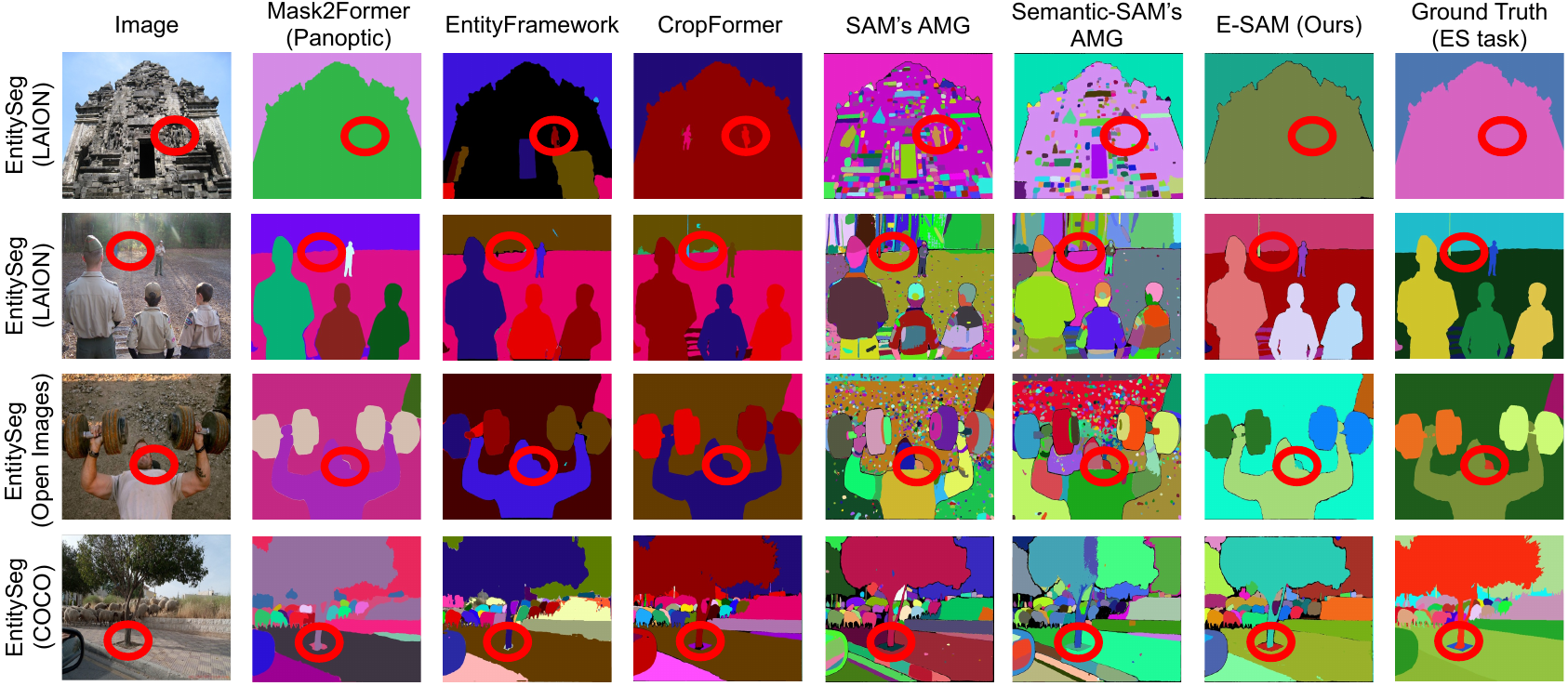}
    \vspace{-3pt}
    \caption{Example visualizations comparing various methods on different data sources in EntitySeg \cite{qi2022open}, including COCO \cite{lin2014microsoft}, LAION \cite{schuhmann2022laion}, and Open Images \cite{kuznetsova2020open}. The corresponding ground truth of the entity segmentation task is provided in the last column.
    }
    \vspace{-10pt}
    \label{Fig: Visual_entity}
\end{figure*}

\section{Experiments}
\subsection{Datasets and Implementation Details}

\noindent\textbf{Datasets}
For qualitative comparison, we adopt the validation set of the EntitySeg dataset \cite{qi2022high}. The EntitySeg dataset comprises 33,227 high-resolution images aggregated from COCO \cite{lin2014microsoft}, ADE20K\cite{zhou2017scene}, Pascal VOC\cite{everingham2010pascal}, LAION \cite{schuhmann2022laion}, Open Images \cite{kuznetsova2020open}, and many other famous datasets. It features precise, class-agnostic entity masks for open-world segmentation. Unlike prior datasets, EntitySeg emphasizes high-quality annotations and complex scenarios. The validation set of EntitySeg has 1,314 images in total.

To test the robustness of the proposed method, we conduct visual comparison in the paper and the supplementary material with images from SA1B \cite{kirillov2023segment}, T-LESS \cite{hodan2017t}, Urban100 \cite{Huang-CVPR-2015}, V2X-SIM \cite{li2022v2x}, Sketch \cite{eitz2012hdhso}, and other self-collected images from the internet.

\noindent\textbf{Implementation details} For the entire evaluation process, we utilized two NVIDIA A40 GPUs along with eight NVIDIA 3090 GPUs. We adopt the pre-trained SAM backbones, specifically ViT-B, ViT-L, and ViT-H. Moreover, we adopt the $AP^e$ metric for open-world entity segmentation.

\begin{table}[t!]
\small
\centering
\renewcommand{\arraystretch}{1.1}
\setlength{\tabcolsep}{0.4pt} 
\begin{tabular}{c|c|c|ccc}
\toprule
\multirow{2}{*}{} & \multirow{2}{*}{\textbf{Methods}} & \multirow{2}{*}{\textbf{Backbone}} & \multicolumn{3}{c}{\textbf{EntitySeg}} \\ \cline{4-6}
                  &                          &                           & AP$^e$  & AP$^e_{50}$ & AP$^e_{75}$ \\ \hline
\multirow{5}{*}{\makecell{Panoptic \\ Methods}}
    & $\circ$ \ Mask-RCNN~\cite{He2017MaskR}          & Swin-T & 28.4 & 49.2 & 28.1 \\ \cdashline{2-6}[1pt/1pt]
    & $\circ$ \ Mask Transfiner~\cite{Ke2021MaskTF}     & Swin-T & 33.7 & -    & -    \\ \cdashline{2-6}[1pt/1pt]
    & $\circ$ \ PatchDCT ~\cite{Wen2023PatchDCTPR}           & Swin-T & 35.4 & -    & -    \\ \cdashline{2-6}[1pt/1pt]
    & \multirow{2}{*}{$\circ$ \ Mask2Former ~\cite{Cheng2021MaskedattentionMT}}        & Swin-T & 40.9 & 58.1 & 41.6 \\
    &                               & Swin-L & 46.2 & 63.7 & 47.5 \\ \hline

\multirow{3}{*}{ES Methods}
    & $\circ$ \ EntityFramework~\cite{qi2022open}     & FPN    & 29.9 & 47.6 & 30.1 \\ \cdashline{2-6}[1pt/1pt]
    & \multirow{2}{*}{$\circ$ \ CropFormer~\cite{qi2022high}}          & Swin-T & 42.7 & 59.7 & 43.8 \\
    &                               & Swin-L & 48.0 & 65.3 & 49.3 \\ \hline

\multirow{8}{*}{\makecell{SAM-based \\ Methods}}
    & \multirow{3}{*}{$\circ$ \ SAM~\cite{kirillov2023segment} }                & ViT-B  & 13.7    & 19.1    & 13.4    \\
    &  & ViT-L  & 19.7    & 28.8    & 18.9    \\
    &  & ViT-H  &  20.1   &  32.9  &   19.4  \\ \cdashline{2-6}[1pt/1pt]
    & \multirow{2}{*}{$\circ$ \ Semantic-SAM~\cite{li2023semantic}}
    & Swin-T & 16.1    & 22.7    & 16.4  \\
    && Swin-L & 17.0    & 24.6    & 17.2    \\ \cdashline{2-6}[1pt/1pt]
    & \multirow{3}{*}{\textcolor{red}{\ding{80}} \ E-SAM (Ours)} & \colorbox{gray!20} {ViT-B} & \colorbox{gray!20} {40.1} & \colorbox{gray!20} {57.8} & \colorbox{gray!20} {38.9} \\
    & & \colorbox{gray!20} {ViT-L} & \colorbox{gray!20} {45.9} & \colorbox{gray!20} {62.7} & \colorbox{gray!20} {45.3} \\
    & & \colorbox{gray!20} {ViT-H} & \colorbox{gray!20} {\textbf{50.2}} & \colorbox{gray!20}  {\textbf{66.8}} & \colorbox{gray!20} {\textbf{49.9}} \\ \toprule
\end{tabular}
\caption{Comparison with prior methods ( panoptic-based, ES-based, and SAM-based ) on the EntitySeg benchmark.}
\vspace{-10pt}
\label{tab:entity_seg_comparison}
\end{table}

\subsection{Comparisons with Existing Works}
In this work, we conducted experiments on the EntitySeg dataset using both high-resolution (HR) and low-resolution (LR) testing subsets. Tabs.~\ref{tab:entity_seg_comparison} and~\ref{tab:entityseg_lr_comparison} present the comparative results for various existing methods and different backbones. Fig.~\ref{Fig: Visual_entity} demonstrates the visual comparisons.

\noindent\textbf{Comparison with Panoptic Methods.}
Tab.~\ref{tab:entity_seg_comparison} compares our E-SAM with several state-of-the-art (SOTA) panoptic segmentation methods, including Mask-RCNN~\cite{He2017MaskR}, Mask Transfiner~\cite{Ke2021MaskTF}, PatchDCT~\cite{Wen2023PatchDCTPR} and Mask2Former~\cite{Cheng2021MaskedattentionMT}. Our E-SAM with ViT-B and ViT-L achieves comparable performance to Mask2Former. Furthermore, with the ViT-H backbone, our method outperforms all baselines, achieving $AP^e$ scores of \textbf{50.2}, surpassing Mask2Former’s best performance.
The results of the LR subset show that our E-SAM  excels with a score of \textbf{48.9} when using the ViT-H backbone.

\begin{table}[t!]
\small
\centering
\renewcommand{\arraystretch}{1}
\setlength{\tabcolsep}{10pt} 
\begin{tabular}{c: c: c}
\toprule
\textbf{Method} & \textbf{Backbone} & {\makecell{\textbf{EntitySeg-}\\\textbf{LR} ($AP_{L}^e$)}} \\ \hline
\multirow{2}{*}{$\circ$ Mask2Former~\cite{Cheng2021MaskedattentionMT}} & Swin-T & 38.8 \\ \cdashline{2-3}[1pt/1pt]
& Swin-L & 44.4 \\ \hline
\multirow{2}{*}{$\circ$ CropFormer~\cite{qi2022high}}
& Swin-T & 40.6 \\ \cdashline{2-3}[1pt/1pt]
& Swin-L & 45.8 \\ \hline
\multirow{3}{*}{$\circ$ SAM~\cite{kirillov2023segment}} & ViT-B  & 10.5 \\ \cdashline{2-3}[1pt/1pt]
& ViT-L  &   17.3    \\ \cdashline{2-3}[1pt/1pt]
& ViT-H  &  17.0 \\ \hline
\multirow{3}{*}{$\circ$ SAM\textsuperscript{O}~\cite{kirillov2023segment}} & ViT-B  & 12.2 \\ \cdashline{2-3}[1pt/1pt]
& ViT-L  &   21.6    \\ \cdashline{2-3}[1pt/1pt]
& ViT-H  &  23.7 \\ \hline
\multirow{3}{*}{\textcolor{red}{\ding{80}} E-SAM (Ours)} & ViT-B  & \colorbox{gray!20} {35.8} \\ \cdashline{2-3}[1pt/1pt]
& ViT-L  &  \colorbox{gray!20} {43.6}    \\ \cdashline{2-3}[1pt/1pt]
& ViT-H  &  \colorbox{gray!20}{\textbf{48.9}}  \\ \toprule
\end{tabular}
\caption{Comparison of various entity segmentation approaches with different backbones on the EntitySeg-LR benchmark, evaluated in terms of $AP_{L}^e$. SAM\textsuperscript{O} denotes object-level map.}
\vspace{-12pt}
\label{tab:entityseg_lr_comparison}
\end{table}

\begin{figure*}[t!]
    \centering
    \includegraphics[width=0.9\textwidth]{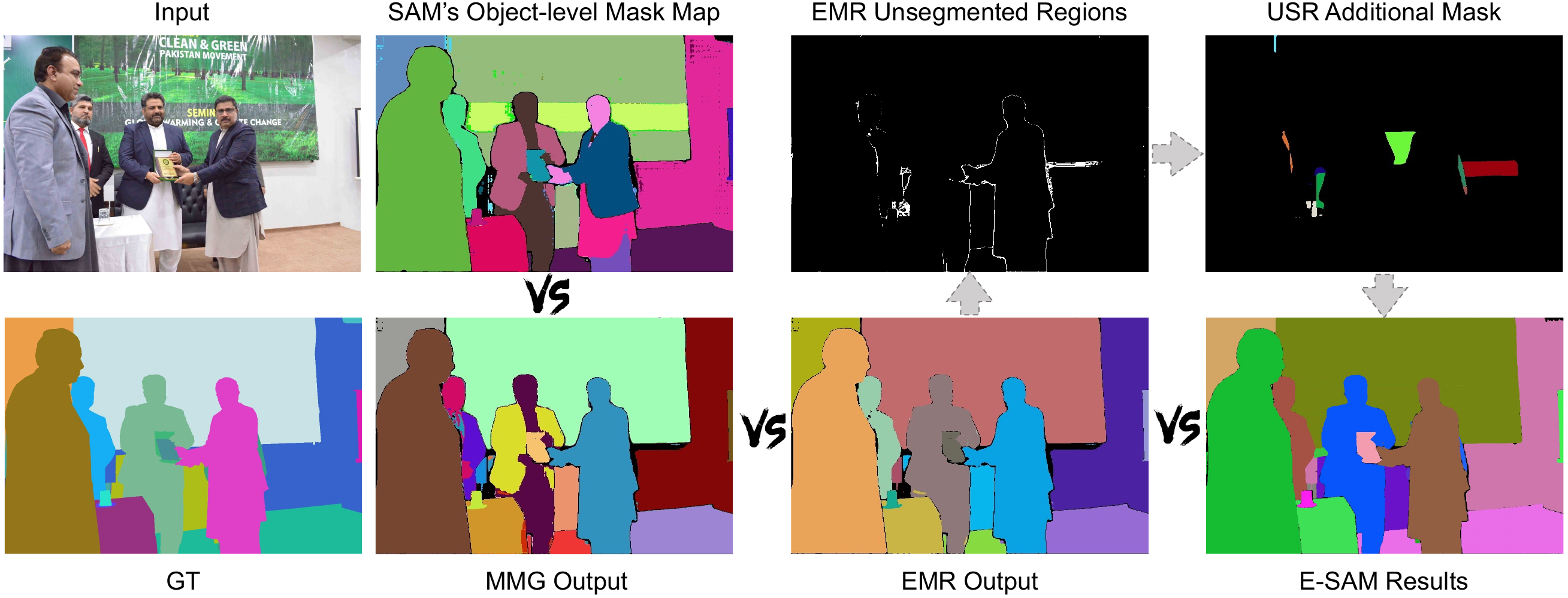}
    \vspace{-5pt}
    \caption{Example visualizations of the sequential application of the three key modules.
    }
    \vspace{-14pt}
    \label{Fig: visual_Alation}
\end{figure*}

\noindent\textbf{Comparison with ES Methods}
In the second group, we compare our E-SAM with the methods specifically designed for ES task, such as EntityFramework~\cite{qi2022open} and CropFormer~\cite{qi2022high}. These methods are tailored to entity segmentation, giving them an advantage with similarly sized backbones. Although our E-SAM shows slightly lower performance compared to CropFormer when using the ViT-B and ViT-L backbones. However, with larger backbones, such as ViT-H, our E-SAM achieves a notable improvement, reaching \textbf{50.2} $AP^e$ and \textbf{48.9 }$AP_L^e$ without incurring additional training costs associated with larger model training.

\begin{table}[t!]
\small
\centering
\renewcommand{\arraystretch}{1.1}
\setlength{\tabcolsep}{8pt} 
\begin{tabular}{|c|c|c|ccc|}
\toprule
\multirow{2}{*}{MMG} & \multirow{2}{*}{EMR} & \multirow{2}{*}{\makecell{USR}} & \multicolumn{3}{c|}{EntitySeg} \\ \cline{4-6}
&                               &                                           & $AP_{L}^e$ & $AP_L^{e_{50}}$ & $AP_L^{e_{75}}$ \\ \hline
                     -& -& - & 17.3 &23.1 & 18.0 \\ \hline
\checkmark & -& - & 20.3 &40.4 & 19.7 \\ \hline
-& \checkmark & - & 29.5 &50.6 & 29.1 \\ \hline
-& - & \checkmark & 22.7 & 31.4 & 23.0 \\ \hline
- &\checkmark & \checkmark & 40.8 & 57.4 & 40.1 \\ \hline
\checkmark & - & \checkmark & 38.1 &54.9 & 37.2 \\ \hline
\checkmark & \checkmark  & -& 35.0 & 54.1 & 35.4 \\ \hline
\checkmark & \checkmark  & \checkmark & \colorbox{gray!20} {\textbf{43.6}} &\colorbox{gray!20} {\textbf{60.8}} & \colorbox{gray!20} {\textbf{43.1}} \\ \toprule
\end{tabular}
\caption{Ablation study on three key modules of E-SAM.}
\vspace{-15pt}
\label{tab:Ablation}
\end{table}

\noindent\textbf{Comparison with SAM-based Methods}
In the third comparison, we evaluate our E-SAM alongside other SAM-based methods, specifically SAM~\cite{kirillov2023segment} and Semantic-SAM~\cite{li2023semantic}.
In the high-resolution setting, SAM (ViT-H) achieves only an $AP^e$ of \textbf{20.1}, whereas our E-SAM, utilizing the ViT-H backbone, significantly improves this score to \textbf{50.2} $AP^e$. Similarly, compared to Semantic-SAM (Swin-L), which scores \textbf{17.0} $AP^e$, our E-SAM with the ViT-L backbone surpasses it by a margin of \textbf{+28.9} $AP^e$.
For low-resolution images, SAM (ViT-H) achieves an $AP_L^e$ of \textbf{23.7}, while our E-SAM with the ViT-H backbone attains \textbf{48.9} $AP_L^e$, improving the performance by \textbf{+25.2} $AP_L^e$.

\subsection{Ablation Studies and Analysis}

\subsubsection{Ablation of Three Key Module}

\noindent\textbf{Effectiveness of MMG Module.}
Tab.~\ref{tab:Ablation} highlights the effectiveness of the MMG module based on ViT-L backbone and EntitySeg-LR test set. Adding MMG to the baseline (which consists solely of AMG mode) results in a notable performance increase, with $AP_L^e$ improving from \textbf{17.3} to \textbf{20.3}. Furthermore, even with EMR and USR applied, incorporating MMG significantly improves performance, raising $AP_L^e$ from \textbf{40.8} to \textbf{43.6}. This emphasizes the crucial role of MMG in generating refined object-level masks, leading to enhanced ES performance. Fig.~\ref{Fig: visual_Alation} also visually illustrates how MMG refines SAM's object-level masks, such as for the person on the right and the painting on the wall.

\noindent\textbf{Effectiveness of EMR Module.}
The addition of the EMR module significantly improves performance over both the baseline and the MMG-only setup (+\textbf{12.2} and +\textbf{14.7}, respectively). When EMR is incorporated alongside MMG and USR, performance improves notably from \textbf{38.1} to \textbf{43.6} in $AP_L^e$, emphasizing EMR's effectiveness in refining overlapping masks and enhancing the accuracy of entity-level segmentation. Fig.~\ref{Fig: visual_Alation} also demonstrates that, with the help of EMR, the identical person in the MMG output is effectively fused into a single entity mask.

\begin{table}[t!]
\footnotesize
\centering
\renewcommand{\arraystretch}{0.5}
\setlength{\tabcolsep}{10pt}
\begin{tabular}{cc:cc:cc}
\toprule
\multicolumn{2}{c}{$\theta_O$} & \multicolumn{2}{c}{$\gamma_O$} & \multicolumn{2}{c}{$\delta$}      \\
\cmidrule(lr){1-2} \cmidrule(lr){3-4} \cmidrule(lr){5-6}
Value      & AP$^e$            & Value      & AP$^e$            & Value        & AP$^e$             \\ \cdashedline{1-2} \cdashedline{3-4} \cdashedline{5-6}
0.7        & 19.5              & 0.3        & 17.6              & 0.01         & 34.1               \\ \cdashedline{1-2} \cdashedline{3-4} \cdashedline{5-6}
0.75       & 19.8              & 0.9        & 18.2              & 0.05         & \textbf{35.0}      \\ \cdashedline{1-2} \cdashedline{3-4} \cdashedline{5-6}
0.8        & \textbf{20.3}     & 0.6        & \textbf{20.3}     & 0.1          & 34.4               \\ \cdashedline{1-2} \cdashedline{3-4} \cdashedline{5-6}
0.9        & 19.6              & 1.0        & 19.6              & 0.2          & 33.8               \\ \hline \hline
\multicolumn{2}{c}{$\tau$}     & \multicolumn{2}{c}{$\rho$}     & \multicolumn{2}{c}{Point Prompts} \\ \cmidrule(lr){1-2} \cmidrule(lr){3-4} \cmidrule(lr){5-6}
Value      & AP$^e$            & Value      & AP$^e$            & Value        & AP$^e$             \\ \cdashedline{1-2} \cdashedline{3-4} \cdashedline{5-6}
0          & 34.1              & 0          & 43.4              & 16/16        & 39.8               \\ \cdashedline{1-2} \cdashedline{3-4} \cdashedline{5-6}
0.05       & 34.5              & 0.1        & \textbf{43.6}     & 16/32        & 41.9               \\ \cdashedline{1-2} \cdashedline{3-4} \cdashedline{5-6}
0.1        & \textbf{35.0}     & 0.3        & 42.9              & 16/64        & 42.5               \\ \cdashedline{1-2} \cdashedline{3-4} \cdashedline{5-6}
0.2        & 34.2              & 0.5        & 42.6              & 32/64        & \textbf{43.6}      \\ \toprule
\end{tabular}

\caption{Ablation study on hyperparameters and the number of point prompts. All performance is evaluated in each module.}
\vspace{-20pt}
\label{tab:Hyperparameter Ablation}
\end{table}

\begin{figure*}[t!]
    \centering
    \includegraphics[width=0.95\textwidth]{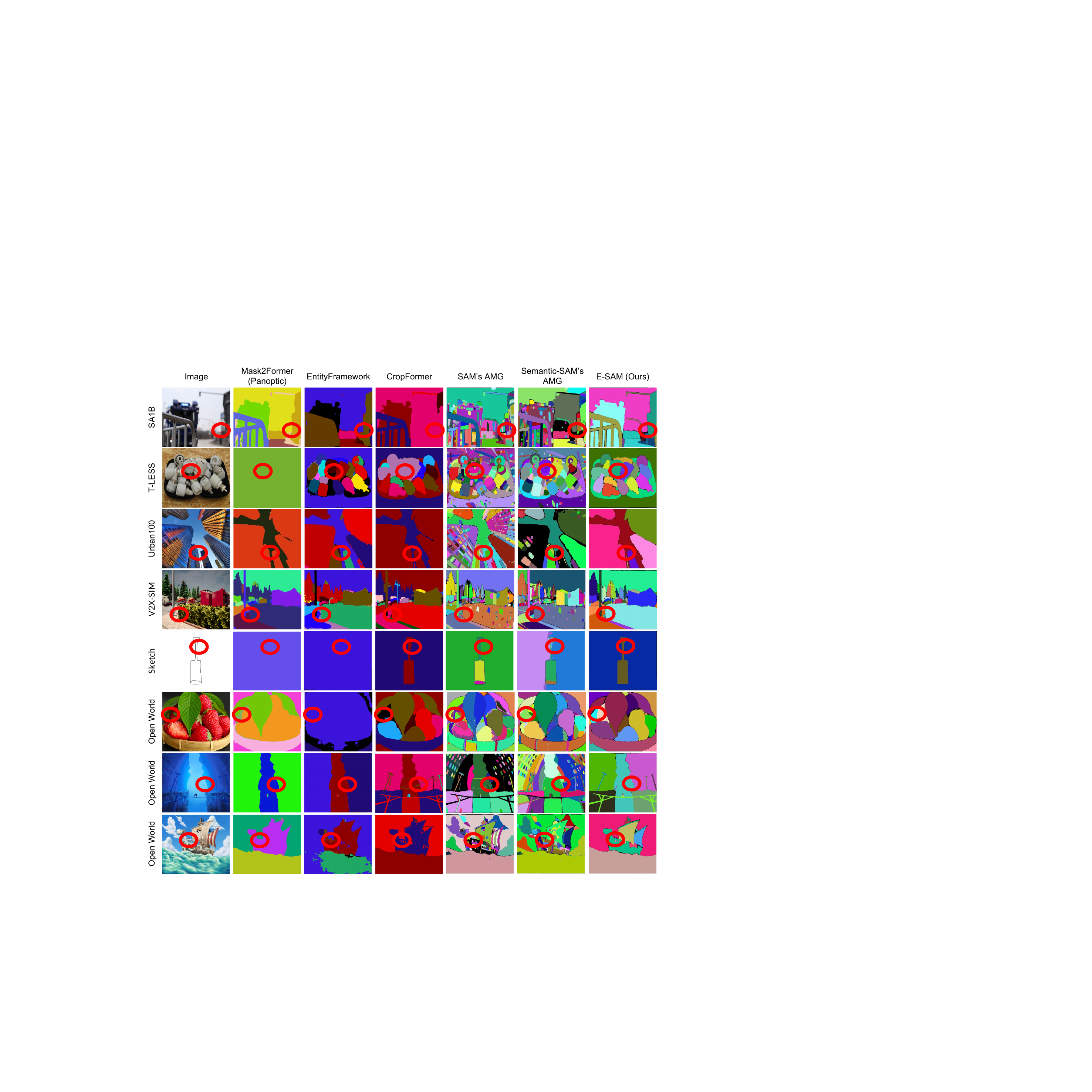}
    \vspace{-3pt}
    \caption{Example visualizations comparing various methods on SA1B \cite{kirillov2023segment}, T-LESS \cite{hodan2017t}, Urban100 \cite{Huang-CVPR-2015}, V2X-SIM \cite{li2022v2x}, Sketch \cite{eitz2012hdhso}, and other open-world images.
    }
    \vspace{-12pt}
    \label{Fig: Robust_Comparsion}
\end{figure*}

\noindent\textbf{Effectiveness of USR Module.}
Incorporating the USR module improves the performance by \textbf{5.4} $AP_L^e$ over the baseline.
When adding USR into MMG and EMR modules separately, the performance improves by \textbf{17.8} and \textbf{11.3} in $AP_L^e$, respectively.
Additionally, adding USR to the MMG and EMR combination further improves $AP_L^e$ from \textbf{35.0} to \textbf{43.6}, emphasizing USR's ability to refine under-segmented areas effectively. Fig.~\ref{Fig: visual_Alation} also demonstrates the EMR's unsegmented regions and the USR's additional masks.

\noindent\textbf{Ablation study of Hyperparameters.}
Tab. \ref{tab:Hyperparameter Ablation} presents the ablation study on hyperparameters and point prompts. Through extensive experiments on the EntitySeg-LR dataset, we carefully calibrated $\theta_O$ and $\gamma_O$ using the MMG module, while $\delta$ and $\tau$ were optimized via MMG and EMR. The final tuning of $\rho$ and point-prompts-per-side was conducted with the complete E-SAM framework, achieving optimal performance and demonstrating superior effectiveness in our experimental settings. The optimal values for each hyperparameter are as follows: $\theta_O$ achieves the best performance at 0.8, $\gamma_O$ at 0.6, and $\delta$ at 0.05. The optimal value for $\tau$ is 0.1, while $\rho$ is 0.1. For the number of point prompts, the best performance is obtained with 32/64, yielding an $AP^e$ of 43.6.

\begin{figure*}[t!]
    \centering
    \includegraphics[width=0.95\textwidth]{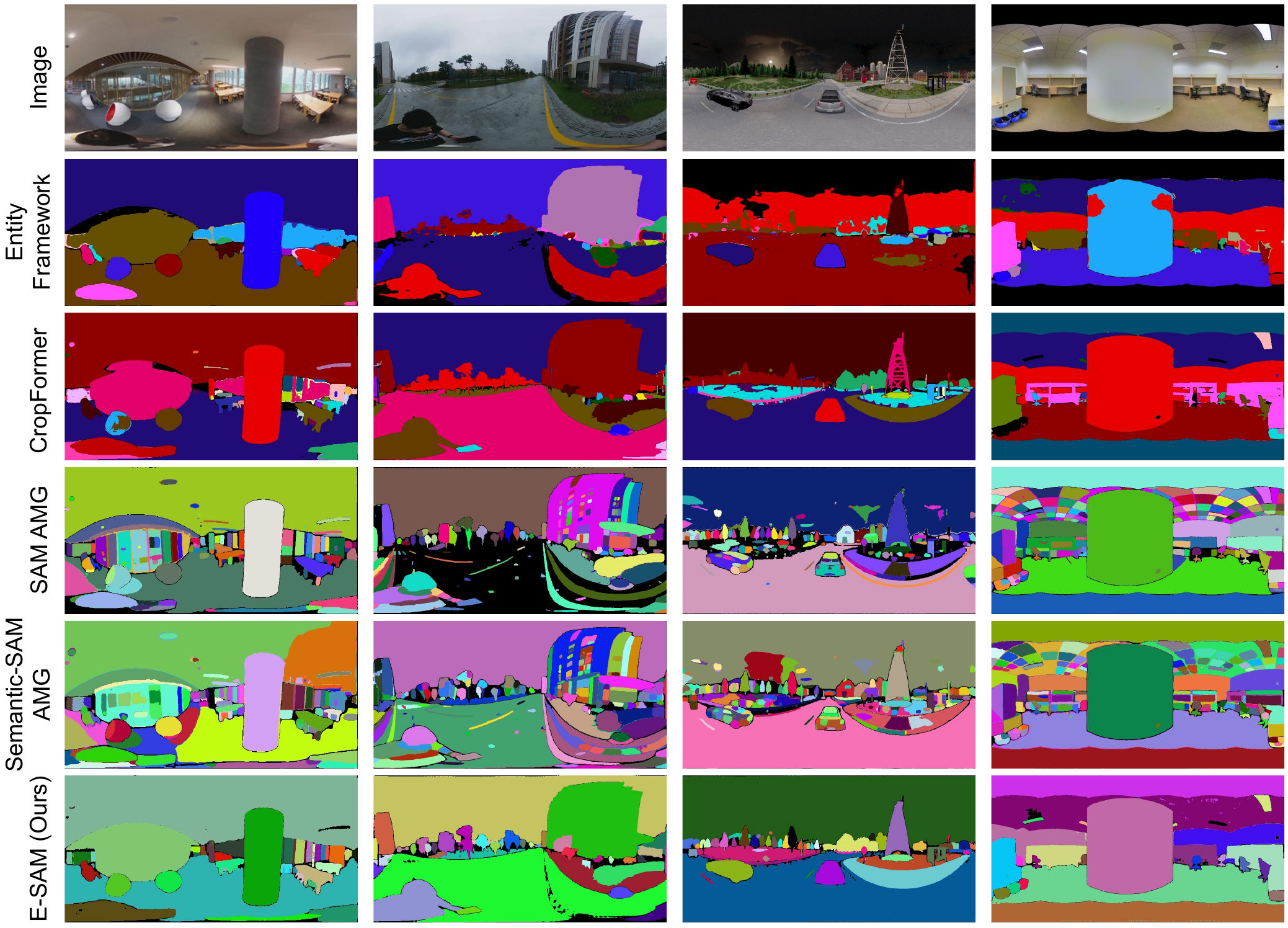}
    \caption{Segmentation visualization of various approaches for the 360 images in indoor\&outdoor environment (self-collected), simulation \cite{zhang2024behind}, and Stanford2D3D \cite{armeni2017joint} dataset.
    }
    \vspace{-15pt}
    \label{Fig: Visual_360supp_2}
\end{figure*}

\subsection{Other Analysis}

\noindent\textbf{Performance of Object-Level Mask from SAM.}
In this section, we discuss the performance of SAM's object-level masks (See Tab.~\ref{tab:entityseg_lr_comparison}). When SAM returns only the object-level mask map, the reduction in overlapping part- and subpart-level masks improves $AP_L^e$ scores from \textbf{17.0} to \textbf{23.7}. This demonstrates the feasibility of converting SAM's object-level masks into ES masks. Notably, our E-SAM consistently surpasses SAM$^O$ across all backbones, underscoring its novelty and effectiveness.

\begin{table}[t!]
    \footnotesize
    \centering
    \renewcommand{\arraystretch}{1.2}
    \setlength{\tabcolsep}{1.5pt} 
    \begin{tabular}{c:c:c:c}
        \toprule
        \textbf{Method} & \textbf{Backbone} & \textbf{Params (M)}  & \textbf{Inference (s)}\\ \hline
        \multirow{1}{*}{$\circ$ Mask2Former ~\cite{Cheng2021MaskedattentionMT}} & Swin-L & 234  & 0.26\\ \hline
        \multirow{1}{*}{$\circ$ EntityFramework ~\cite{qi2022open}} & FPN & - & 0.34\\ \hline
        \multirow{1}{*}{$\circ$ CropFormer ~\cite{qi2022high}} & Swin-L & 234  & 1.08\\ \hline
        \multirow{4}{*}{\textcolor{red}{\ding{80}} E-SAM (Ours)} & \textbf{ViT-H} & 636  & \textbf{9.84}\\ \cdashline{2-4}[1pt/1pt]
        & SAM (Encoder) & 636  & 5.89\\ \cdashline{2-4}[1pt/1pt]
        & MMG & -  & 1.22\\ \cdashline{2-4}[1pt/1pt]
        & EMR & -  & 1.58\\ \cdashline{2-4}[1pt/1pt]
        & USR & -  & 1.15\\ \toprule
    \end{tabular}
    \vspace{-5pt}
        \caption{Comparison of model complexity and inference time across different approaches.}
    \label{tab:inference time}
\vspace{-17pt}
\end{table}

\noindent\textbf{Inference Time Comparision.} In this section, we discuss the inference time of E-SAM. As shown in Tab. \ref{tab:inference time}, E-SAM takes \textbf{9.84} seconds on average to process a high-resolution image with the ViT-H backbone, which is higher than existing ES methods. However, this trade-off is justified by the avoidance of substantial training costs, with E-SAM introducing efficient post-processing operations. Notably, the SAM encoder accounts for \textbf{5.89} seconds, while each proposed module operates around \textbf{1} second. Given the strong generalization capabilities demonstrated in Fig. \ref{Fig: Robust_Comparsion}, the additional inference time is warranted.

\noindent\textbf{Superpixels Method Discussion.}
Although our E-SAM leverages superpixels in all three modules, its novelty does not stem from a simple combination of superpixels and SAM. In the supplementary material, we compare different superpixel generation methods between SLIC~\cite{achanta2012slic} and Felzenszwalb~\cite{Felzenszwalb2004EfficientGI} and further evaluate the direct integration of superpixels with SAM for entity segmentation.



\noindent \textbf{Comparisons on Simulation Images.}
As shown in the fourth row of Fig.~\ref{Fig: Robust_Comparsion}, E-SAM outperforms other methods, particularly when handling numerous similar objects in the background. This effectively demonstrates the robustness and effectiveness of the E-SAM framework. Notably, unlike CropFormer, which struggles with distinguishing same-class objects in the background and tends to exaggerate foreground elements, E-SAM achieves a better balance between foreground and background segmentation. For instance, in the third row, E-SAM accurately segments the trees in the background.

\noindent \textbf{Comparisons on Sketches.} E-SAM's performance is close to CropFormer when handling sketches, as shown in the fifth row of Fig. \ref{Fig: Robust_Comparsion}. Since sketches consist of only lines, it confuses SAM and affects the information for E-SAM. In contrast to SAM and Semantic-SAM, our E-SAM effectively removes overlapping masks and successfully merges masks corresponding to the same entity. Given its training-free nature, this highlights the novelty of our E-SAM even more clearly.

\noindent \textbf{Comparisons on Cartoons.} For the last two rows in Fig.~\ref{Fig: Robust_Comparsion}, we conducted a visual comparison between prior SOTA entity segmentation methods and SAM-based methods using open-world cartoon-style images. It can be observed that our E-SAM outperforms all methods except CropFormer. Given that our approach is training-free, achieving comparable performance to CropFormer is an acceptable trade-off, demonstrating the practicality and efficiency of our method.

\noindent\textbf{More Robustness Comparisons.}
In this section, we test the robustness of E-SAM to address concerns about its performance being limited to the EntitySeg dataset. Fig.~\ref{Fig: Robust_Comparsion} demonstrates that E-SAM produces more robust ES results across other segmentation datasets and online open-world images.
To further evaluate the real-world robustness of E-SAM, we tested it on 360 images (indoor/outdoor, real/synthetic).
Fig. \ref{Fig: Visual_360supp_2} demonstrates the segmentation results of various methods on 360 images.
In Fig.~\ref{Fig: Visual_360supp_2}, we compare our E-SAM with prior methods across indoor and outdoor open-world scenes, synthetic environments, and the indoor benchmark dataset Stanford2D3D~\cite{armeni2017joint}. In both indoor and outdoor open-world scenarios, E-SAM achieves accurate segmentation, such as the table in the first column, and demonstrates strong performance even for small objects near the equator of the image, significantly outperforming the other methods. In synthetic 360 environments, E-SAM maintains refined and detailed entity-level segmentation results, showing resilience against real-synthetic domain gaps. In the Stanford2D3D dataset, E-SAM consistently delivers precise masks for distant or smaller objects, while our three modules collaboratively reduce overlapping masks and effectively refine SAM AMG results into entity-level segmentation maps, outperforming SAM and Semantic-SAM.

\section{Conclusion and Future Work}

In this paper, we introduced a novel framework -- E-SAM that enhances SAM's AMG for entity segmentation without additional training overhead. Our framework integrates MMG for refining multi-level mask generation, EMR for addressing overlapping segments and fusing entity-level masks, and USR for refining under-segmented regions. Extensive comparative experiments and ablation studies demonstrate that E-SAM achieves state-of-the-art performance and validates the effectiveness of each module.

\noindent\textbf{Future Work.} Future work includes fine-tuning E-SAM to further boost its performance, exploring lightweight strategies to accelerate inference for real-time entity segmentation, and investigating knowledge transfer methods to more compact models for broader applicability.

{
    \small
    \bibliographystyle{ieeenat_fullname}
    \bibliography{main}
}

\end{document}